\documentclass[11pt]{article}

\usepackage[final]{acl}

\usepackage{times}
\usepackage{latexsym}

\usepackage[T1]{fontenc}

\usepackage[utf8]{inputenc}

\usepackage{microtype}

\usepackage{inconsolata}

\usepackage{graphicx}

\usepackage{amsmath}
\usepackage{amsfonts}
\usepackage{graphicx}
\usepackage{booktabs}
\usepackage{multirow}

\title{Enhancing Video Representations with Spatiotemporal-Semantic Residual to Mitigate Hallucinations in Video Large Multimodal Models}

\author{
 \textbf{Yuansheng Gao\textsuperscript{1,$\dagger$,$\ddagger$}},
 \textbf{Jinman Zhao\textsuperscript{2,$\ddagger$}},
 \textbf{Tong Zhang\textsuperscript{1}},
 \textbf{Xingguo Xu\textsuperscript{3}},
 \textbf{Wenbin Xing\textsuperscript{4}},
\\
 \textbf{Han Bao\textsuperscript{1,$\S$}},
 \textbf{Zonghui Wang\textsuperscript{1,$\S$}},
 \textbf{Wenzhi Chen\textsuperscript{1}}
\\
\\
 \textsuperscript{1}Zhejiang University,
 \textsuperscript{2}University of Toronto,
\\
 \textsuperscript{3}Dalian University of Technology,
 \textsuperscript{4}Sun Yat-sen University
\\
 \small{
   \textsuperscript{$\dagger$}\textbf{Email:} \href{mailto:y.gao@zju.edu.cn}{y.gao@zju.edu.cn}
 }
\\
 \small{
   \textsuperscript{$\ddagger$}Co-first authors
 }
 \small{
   \textsuperscript{$\S$}Corresponding authors
 }
}

\begin{document}
\maketitle
\begin{abstract}
Although Video Large Multimodal Models have achieved strong performance in video understanding, they still suffer from hallucination. Existing inference-time intervention methods usually modify videos under the contrastive decoding framework, but their heuristic designs bring limited improvements and increase inference latency.
To address these issues, we propose ViSSRes, an inference-time intervention method that enhances video representations through a lightweight MLP-style network. Specifically, we use a contrastive random walk approach to characterize the spatiotemporal consistency of video representations, and introduce conditional mutual information to associate video representations with the model’s semantic understanding. With the model backbone kept frozen, ViSSRes learns residuals for video representations and optimizes them from both spatiotemporal and semantic consistency perspectives.
During inference, ViSSRes requires only a single forward pass and introduces no substantial additional inference cost. Experiments show that ViSSRes reduces the hallucination rate of LLaVA-NeXT-Video on EventHallusion by 40.69\% and improves video understanding on MMVU by 18.36\% under the CoT setting, demonstrating its effectiveness in mitigating hallucinations.
\end{abstract}

\section{Introduction}
\begin{figure}
    \raggedright
    \includegraphics[width=0.9\linewidth]{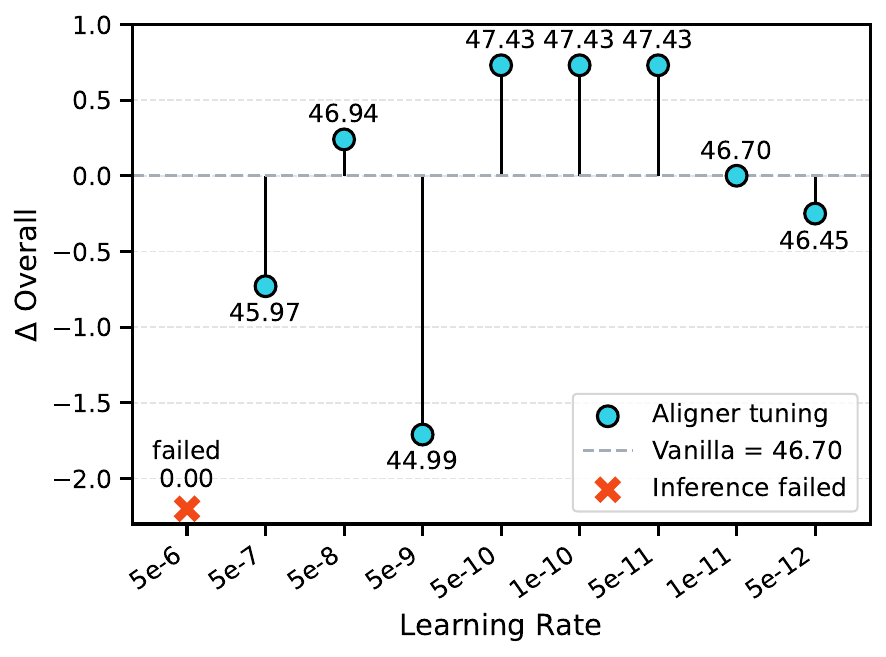}
    \caption{Role of contrastive random walk~\cite{jabri2020space} under different learning rates. Using Video-LLaVA~\cite{zhulanguagebind} on EventHallusion~\cite{zhang2024eventhallusion}, a learning rate of 5e-6 led to excessive spatiotemporal alignment in video representations, resulting in an Overall accuracy of 0.}
    \label{fig:motivation}
\end{figure}

The advancement of Video Large Multimodal Models (VideoLMMs) has greatly accelerated progress in video understanding, question answering, and multimodal reasoning~\cite{xu2024pllava, ma2024vista, wang2025affordance, li2025vidhalluc}. However, hallucination remains a critical issue that undermines their reliability and trustworthiness~\cite{sun2026smartsight, poppi2026countervid, tang2025seeing}. In VideoLMMs, hallucinations typically refer to semantically plausible responses that are inconsistent with the input video content or objective factual evidence~\cite{liu2024tempcompass, zheng2024thinking}. Although hallucination mitigation has been extensively studied in image-based multimodal models~\cite{yue2024less, jiang2025devils, fu2025chip, liumitigating}, relatively fewer efforts have focused on videos, where temporal dynamics introduce additional challenges.

Existing methods for mitigating hallucinations in VideoLMMs can be roughly categorized into preference-based fine-tuning, architectural modification, and inference-time intervention. Preference-based methods~\cite{ding2025pami, huang2025vistadpo} improve response faithfulness through alignment training, but usually require high-quality preference data or substantial training costs. Architectural modification-based methods~\cite{ma2024vista} modify the model structure to enhance cross-modal alignment, which may limit their compatibility and scalability in practical deployment. In contrast, inference-time intervention methods, especially contrastive decoding~\cite{kong2025mhbench, zhang2024eventhallusion, li2025vidhalluc}, provide a lightweight plug-and-play alternative. However, they commonly rely on overly heuristic designs, which limits their robustness across diverse video contents and hallucination types. Moreover, contrastive decoding often requires additional inference passes, increasing the decoding overhead.

To address these limitations, we propose ViSSRes, a lightweight representation optimization method for mitigating hallucinations in VideoLMMs. Instead of relying on manually designed negative samples, ViSSRes optimizes video representations from both spatiotemporal and semantic consistency perspectives. Specifically, we introduce a lightweight spatiotemporal-semantic aligner after the video encoder to refine video representations. For spatiotemporal consistency, we draw inspiration from contrastive random walks~\cite{jabri2020space}, which provide a principled way to characterize temporal coherence in videos. However, directly applying this mechanism only constrains visual-level consistency and lacks explicit alignment with the semantic space. As shown in Figure~\ref{fig:motivation}, optimizing spatiotemporal consistency alone cannot robustly mitigate hallucinations in VideoLMMs. Therefore, we further introduce semantic consistency alignment based on conditional mutual information, encouraging the video representations to remain semantically aligned with the response generated by VideoLMMs.

By jointly optimizing the spatiotemporal and semantic consistency of video representations, ViSSRes mitigates hallucinations while preserving the general capabilities of VideoLMMs. It only trains a lightweight MLP-style aligner and introduces one additional aligner forward pass during inference, resulting in negligible computational overhead. In summary, our main contributions are as follows:
\begin{itemize}
    \item We show that optimizing spatiotemporal consistency alone is insufficient for reliable hallucination mitigation, as excessive spatiotemporal alignment may weaken the compatibility between visual representations and the model's language-semantic space.

    \item We introduce semantic consistency alignment based on conditional mutual information and propose ViSSRes, a lightweight video representation optimization method that jointly preserves spatiotemporal reliability and semantic decodability.

    \item Extensive experiments demonstrate that ViSSRes outperforms existing state-of-the-art (SOTA) inference-time intervention methods. Specifically, it reduces the hallucination rate of LLaVA-NeXT-Video on EventHallusion by 40.69\% and improves video understanding on MMVU by 18.36\% under the CoT setting.
\end{itemize}

\section{Related Work}

\subsection{Video Large Multimodal Models}
In recent years, VideoLMMs have achieved remarkable progress in the field of multimodal understanding across vision and language \cite{bagad2023test, zhangllava}. Early studies were largely extensions of image-language models such as CLIP~\cite{radford2021learning} and BLIP~\cite{li2022blip}, aligning video frames with textual descriptions or questions through frame-level feature aggregation or cross-modal attention mechanisms. With the emergence of Large Language Models (LLMs)~\cite{yuan2025kardia, gao2025mentalmac, yuan2025reversal}, recent works further couple powerful LLM backbones with frozen vision encoders (e.g., CLIP, LanguageBind~\cite{zhulanguagebind}) to extract spatiotemporal visual tokens, yielding instruction-following VideoLMMs capable of open-ended reasoning and dialogue~\cite{gao2023llama, zhang2023video}. These models, such as Video-LLaVA~\cite{lin2024video}, VideoChatGPT~\cite{maaz2024video}, and Valley~\cite{wu2025valley2}, typically process spatiotemporal visual tokens extracted from pretrained vision encoders and inject them into LLMs through lightweight projection layers for multimodal alignment. Despite their strong performance, current VideoLMMs remain prone to hallucination~\cite{zhang2024eventhallusion}. Our work aims to address this issue and enhance the reliability of VideoLMMs.

\subsection{Hallucinations Mitigation for Video Large Multimodal Models}
Hallucination in VideoLMMs typically refers to the phenomenon where generated responses are inconsistent with salient visual evidence or objective facts in the video~\cite{li2025vidhalluc, bae2025mash}. Compared with static images, videos introduce additional temporal dynamics, making hallucination mitigation more challenging~\cite{wang2024videohallucer}.

For VideoLMMs, preference-based methods such as PaMi-VDPO~\cite{ding2025pami} and VistaDPO~\cite{huang2025vistadpo} mitigate hallucinations through preference-based fine-tuning, but they usually require either annotated preference data or substantial training costs. VISTA-LLAMA~\cite{ma2024vista} modifies the attention mechanism to maintain consistent visual influence during generation, yet it does not explicitly model causal reasoning. These methods are effective to some extent, but their reliance on architectural modification limits their general applicability.
In contrast, inference-time intervention methods provide a more lightweight alternative. TCD~\cite{zhang2024eventhallusion} and MotionCD~\cite{kong2025mhbench} construct negative samples by reversing videos or extracting frames, aiming to mitigate hallucinations by reducing spatiotemporal misalignment. However, these heuristic designs are difficult to generalize to diverse hallucination types. DINO-HEAL~\cite{li2025vidhalluc} corrects model attention with DINOv2~\cite{oquab2023dinov2}, introducing little additional overhead but offering limited hallucination mitigation. SmartSight~\cite{sun2026smartsight} filters low-hallucination outputs by generating and evaluating multiple candidate responses, which inevitably increases inference computation.

Overall, existing methods face a trade-off between effectiveness, general applicability and computational efficiency. Unlike these approaches, ViSSRes trains only a lightweight MLP-style network. During inference, it introduces only one additional forward pass of the network, achieving hallucination mitigation with negligible inference overhead.

\section{Preliminary}

Before introducing ViSSRes, we briefly review the general formulation of VideoLMMs and the autoregressive generation mechanism commonly employed during inference.

A typical VideoLMM consists of a video encoder, a multimodal projector, and a LLM. Given an input video $V = \{v_1, v_2, \dots, v_T\}$ consisting of $T$ frames, the video encoder $\mathcal{E}_v(\cdot)$ extracts spatiotemporal video representations:
\begin{equation}
    \mathbf{H}_v = \mathcal{E}_v(V) \in \mathbb{R}^{T \times N \times d},
\end{equation}
where $N$ denotes the number of visual tokens per frame and $d$ is the feature dimension. These video representations are then mapped into the language embedding space through a projection $\mathcal{P}(\cdot)$:
\begin{equation}
    \mathbf{Z}_v = \mathcal{P}(\mathbf{H}_v).
\end{equation}
The visual tokens $\mathbf{Z}_v$ are concatenated with textual tokens and fed into the LLM, which performs multimodal reasoning and generates textual outputs in an autoregressive manner.

Given the visual tokens $\mathbf{Z}_v$ and a prompt $\mathbf{X} = \{x_1, \dots, x_m\}$, the LLM defines a conditional probability
distribution over a response sequence $\mathbf{Y} = \{y_1, \dots, y_L\}$ in an autoregressive manner:
\begin{equation}
    p_\theta(\mathbf{Y} \mid \mathbf{Z}_v, \mathbf{X})
    = \prod_{t=1}^{L} p_\theta(y_t \mid \mathbf{Z}_v, \mathbf{X}, y_{<t}),
\end{equation}
where $p_\theta(\cdot)$ denotes the conditional distribution over text tokens induced by the VideoLMM parameterized by $\theta$ given multimodal inputs.

\section{Methodology}
\label{sec:method}

\begin{figure*}
  \centering
  \includegraphics[width=\linewidth]{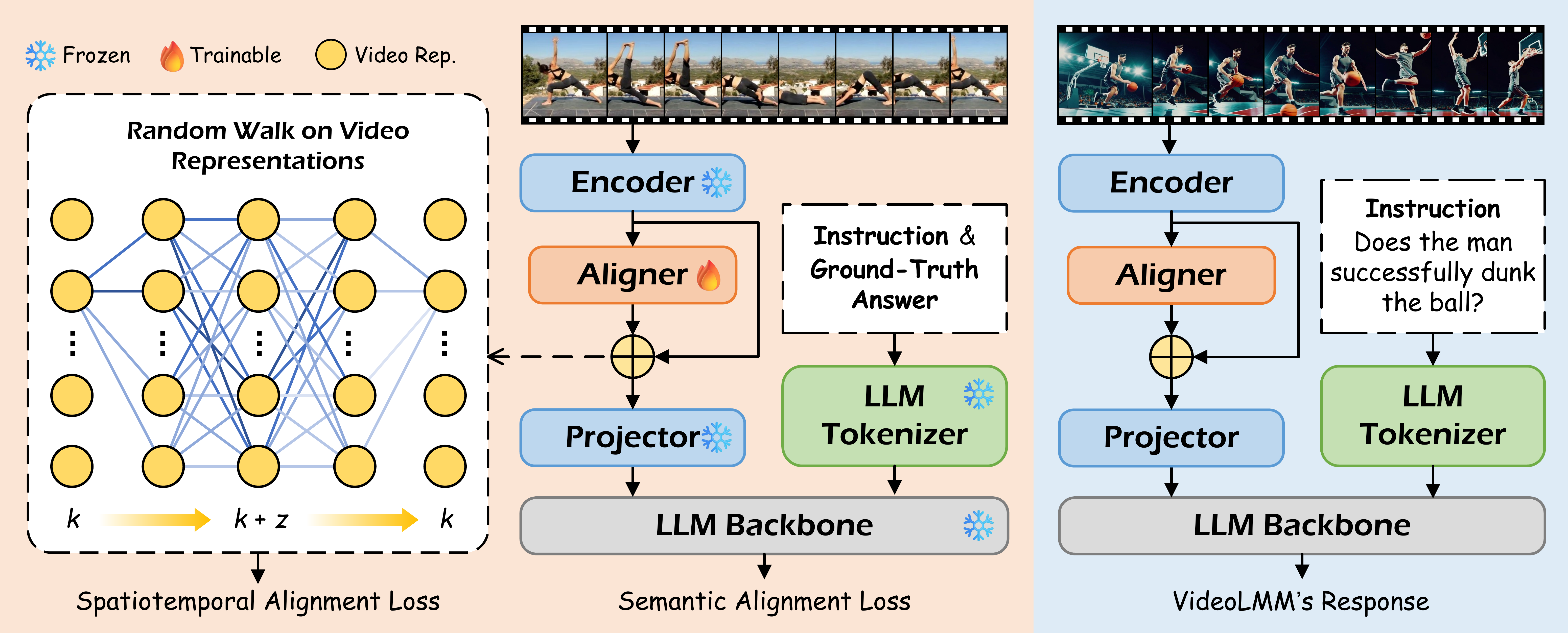}
  \caption{Overview of the proposed ViSSRes. \textbf{Left:} We freeze the VideoLMM and only train the spatiotemporal-semantic aligner. \textbf{Right:} During inference, we insert the aligner after the encoder to reduce hallucinations by optimizing the video representations.}
  \label{fig:overview}
\end{figure*}

To mitigate hallucinations in VideoLMMs, we propose ViSSRes (see Figure~\ref{fig:overview}), a novel inference-time intervention method that enhances video representations from the perspectives of spatiotemporal and semantic consistency alignment. In practice, we design a spatiotemporal-semantic aligner $\mathcal{M}$ to optimize video representations in a residual manner. Formally, this process can be expressed as:
\begin{equation}
\mathbf{H}^{*}_v
= \mathbf{H}_v + \mathcal{M}(\mathbf{H}_v),
\label{eq:aligner}
\end{equation}
where $\mathbf{H}^{*}_v$ denotes the optimized video representations. Subsequent sections detail the training procedure of $\mathcal{M}$ and its role during inference.

\subsection{Spatiotemporal Consistency Alignment}

The multi-frame nature of videos can induce spatiotemporal misalignment in video representations, exacerbating hallucinations in VideoLMMs~\cite{huang2026distorted, wu2025season}. While prior methods mitigate this issue through heuristic contrastive decoding~\cite{kong2025mhbench, zhang2024eventhallusion}, their generalizability remains limited. Contrastive random walk~\cite{jabri2020space}, originally proposed for label propagation, provides a principled way to characterize spatiotemporal consistency. We therefore incorporate it into VideoLMMs to align video representations along the spatiotemporal dimension.

Formally, we model the video representations as a spatiotemporal graph, where each video representation $h_k^n \in \mathbf{H}^{*}_v$ corresponds to a node, with $k$ denoting the frame index and $n$ denoting the feature index within the frame. Edges connect only features in consecutive frames. Affinity between features in consecutive frames is defined via dot-product similarity:
\begin{equation}
S_k^{k+1}(i,j) = \langle h_k^i, h_{k+1}^j \rangle.
\end{equation}
A row-wise softmax produces the transition matrix:
\begin{equation}
\mathcal{T}_{k \to k+1}(i,j)
=\frac{\exp(S_k^{k+1}(i,j)/\tau)}
     {\sum_{n=1}^{N} \exp(S_k^{k+1}(i,n)/\tau)},
\label{eq:transition_matrix}
\end{equation}
where $\tau$ is a temperature parameter that controls the sharpness of the transition distribution.

Following~\citet{jabri2020space}, we apply a random-walk mechanism on the spatiotemporal graph to model long-range consistency while enhancing cross-frame spatiotemporal coherence. Let \(W_k\) denote the state of a random walker at frame \(k\). Since \(\mathcal{T}_{k \to k+1}(i,j) = P(W_{k+1} = j \mid W_k = i)\) specifies the one-step transition probability from node \(h_k^i\) to node \(h_{k+1}^j\), long-range temporal consistency can be modeled by composing multiple such
transitions. By the first-order Markov assumption, the multi-step transition probability from frame \(k\) to frame \(k{+}z\) can be computed as:
\begin{equation}
\begin{aligned}
    \bar{\mathcal{T}}_{k \to k+z}
&= \prod_{i=0}^{z-1} \mathcal{T}_{k+i \to k+i+1}\\
&= P(W_{k+z} \mid W_k).
\end{aligned}
\end{equation}
Thus, \(\bar{\mathcal{T}}_{k \to k+z}\) represents the probability of propagating consistency from frame \(k\) to frame \(k{+}z\).

For a given span \(z\), the forward random walk from frame \(k\) to frame \(k{+}z\) is governed by \(\bar{\mathcal{T}}_{k\to k+z}\), while the corresponding reverse walk along the palindromic sequence is governed by \((\bar{\mathcal{T}}_{k\to k+z})^\top\).
By sequentially composing the forward and reverse walks, the resulting \(2z\)-step round-trip transition from frame \(k\) back to itself is defined as:
\begin{equation}
P(W_{k+2z}\mid W_k)
=
\bar{\mathcal{T}}_{k\to k+z}\,
(\bar{\mathcal{T}}_{k\to k+z})^\top.
\end{equation}
The diagonal entry of this round-trip transition matrix, \(P(W_{k+2z}=i\mid W_k=i)\), corresponds to the probability that a walker starting at feature \(i\) returns to the same token after the \(2z\)-step palindromic walk. 
Based on the above analysis, to deliberately optimize cross-frame spatiotemporal consistency, we define a cycle-consistency score as:
\begin{equation}
\begin{aligned}
c_{k,z}
&=\sum_{i=1}^N 
\log P(W_{k+2z}=i\mid W_k=i)\\
&=
\sum_{i=1}^N 
\log \Big(
\big[
\bar{\mathcal{T}}_{k\to k+z}
(\bar{\mathcal{T}}_{k\to k+z})^\top
\big]_{ii}
\Big).
\end{aligned}
\end{equation}

It is clear that a higher cycle-consistency score indicates a greater chance of the walker returning to its start, reflecting stronger spatiotemporal consistency in the video representation.

Subsequently, we compute the cycle-consistency score over all admissible temporal spans \(\mathcal{Z}=\{1,\dots,Z_{\max}\}\) to optimize the spatiotemporal consistency of video representations, where \(Z_{\max}=T-2\) denotes the largest span for which the palindromic sequence remains entirely within the video. 
Specifically, for a given span \(z\in\mathcal{Z}\), the corresponding set of valid starting indices is \(\mathcal{K}_z=\{\,k \mid k+2z \le T\,\}\). The spatiotemporal loss \(\mathcal{L}_{\mathrm{T}}\) is defined as the average of the negative cycle-consistency scores computed over all temporal spans and their valid starting indices:
\begin{equation} 
\mathcal{L}_{\mathrm{T}} = \frac{1}{\sum_{z\in\mathcal{Z}}|\mathcal{K}_z|} \sum_{z\in\mathcal{Z}} \sum_{k\in\mathcal{K}_z} -c_{k,z}. 
\end{equation}

This objective explicitly optimizes video representations via contrastive random walks, which can mitigate hallucinations from the perspective of aligning spatiotemporal consistency.

\subsection{Semantic Consistency Alignment}
Although contrastive random walk provides a reasonable perspective for optimizing the spatiotemporal consistency of video representations, directly applying it to VideoLMMs remains insufficient. This is because it models consistency only within the video modality, without explicitly constraining the semantic correspondence between video representations and the language space. As shown in Figure~\ref{fig:motivation}, this may cause the optimized representations to overfit spatiotemporal consistency while drifting away from the LLM semantic space. To address this issue, we further introduce semantic consistency alignment as an additional optimization guidance, ensuring that video representations remain semantically aligned with the language space while enhancing spatiotemporal consistency.

To align video representations with semantics, we maximize the conditional mutual information between the video representations $\mathbf{H}^{*}_v$ and the response $\mathbf{Y}$ given the input $\mathbf{X}$:
\begin{equation}
    \max_{\mathbf{H}^{*}_v} I(\mathbf{H}^{*}_v;\mathbf{Y}\mid \mathbf{X})
    \label{eq:o_cmi}.
\end{equation}

This objective directly aligns the video representations with the model's responses, encouraging the representations to capture more information relevant to the answers and thereby reducing potential hallucinations. However, directly optimizing this objective is very challenging, primarily due to the lack of alignment between modalities. Therefore, we leverage the projector $\mathcal{P}$ in VideoLMMs to map visual representations into the embedding space, i.e., $\mathbf{Z}^{*}_v = \mathcal{P}\big(\mathbf{H}^{*}_v\big)$, making Equation~\ref{eq:o_cmi} equivalently expressed as:
\begin{equation}
    \max_{\mathbf{Z}^{*}_v} I(\mathbf{Z}^{*}_v;\mathbf{Y}\mid \mathbf{X}).
    \label{eq:p_cmi}
\end{equation}

According to the definition of conditional mutual information, Equation~\ref{eq:p_cmi} can be expressed in probabilistic form. Since the true conditional distribution is generally inaccessible in practice, we approximate it using the autoregressive response distribution parameterized by the VideoLMM:
\begin{equation}
    \max_{\mathbf{Z}^{*}_v} \log \frac{p_\theta \big(\mathbf{Z}^{*}_v,\mathbf{Y}|\mathbf{X}\big)}{p_\theta \big(\mathbf{Z}^{*}_v|\mathbf{X}\big)p_\theta\big(\mathbf{Y}|\mathbf{X}\big)}.
    \label{eq:h_cmi}
\end{equation}
Although this formulation clarifies the optimization target, it is still not directly tractable. Therefore, we further apply Bayes' theorem to transform it into a computable form:
\begin{equation}
\begin{aligned}
    &\max_{\mathbf{Z}^{*}_v} \log \frac{p_\theta \big(\mathbf{Z}^{*}_v,\mathbf{Y}|\mathbf{X}\big)}{p_\theta \big(\mathbf{Z}^{*}_v|\mathbf{X}\big)p_\theta\big(\mathbf{Y}|\mathbf{X}\big)}\\
    =&\max_{\mathbf{Z}^{*}_v} \log \frac{p_\theta \big(\mathbf{Z}^{*}_v,\mathbf{Y},\mathbf{X}\big)\big/p_\theta\big(\mathbf{X}\big)}{p_\theta \big(\mathbf{Z}^{*}_v|\mathbf{X}\big)p_\theta\big(\mathbf{Y}|\mathbf{X}\big)}\\
    =&\max_{\mathbf{Z}^{*}_v} \log \frac{p_\theta \big(\mathbf{Z}^{*}_v,\mathbf{Y},\mathbf{X}\big)}{p_\theta\big(\mathbf{X}\big)p_\theta \big(\mathbf{Z}^{*}_v|\mathbf{X}\big)p_\theta\big(\mathbf{Y}|\mathbf{X}\big)}\\
    =&\max_{\mathbf{Z}^{*}_v} \log \frac{p_\theta \big(\mathbf{Z}^{*}_v,\mathbf{Y},\mathbf{X}\big)}{p_\theta \big(\mathbf{Z}^{*}_v,\mathbf{X}\big)p_\theta\big(\mathbf{Y}|\mathbf{X}\big)}\\
    =&\max_{\mathbf{Z}^{*}_v} \log \frac{p_\theta \big(\mathbf{Y}\mid\mathbf{Z}^{*}_v,\mathbf{X}\big)}{p_\theta\big(\mathbf{Y}|\mathbf{X}\big)}.
    \label{eq:c_cmi}
\end{aligned}
\end{equation}

At this point, we have derived the objective in Equation~\ref{eq:c_cmi} into a computable form. We further express the above as a minimization problem:
\begin{equation}
\begin{aligned}
    &\min_{\mathbf{Z}^{*}_v} \log \frac{p_\theta\big(\mathbf{Y}|\mathbf{X}\big)}{p_\theta \big(\mathbf{Y}\mid\mathbf{Z}^{*}_v,\mathbf{X}\big)}\\
    =&\min_{\mathbf{Z}^{*}_v} \log p_\theta\big(\mathbf{Y}|\mathbf{X}\big)-\log p_\theta \big(\mathbf{Y}\mid\mathbf{Z}^{*}_v,\mathbf{X}\big).
    \label{eq:min_cmi}
\end{aligned}
\end{equation}

Based on Equation~\ref{eq:min_cmi}, we design the following semantic consistency loss for training:
\begin{equation}
\begin{aligned}
    \mathcal{L}_{\mathrm{S}}
    = \sum_{t=1}^{L}
    \Big[&\log p_{\theta}\big(y_t \mid \mathbf{X}, y_{<t}\big)-\\
    &\log p_{\theta}\big(y_t \mid \mathbf{Z}^*_v, \mathbf{X}, y_{<t}\big)\Big].
    \label{eq:loss_cmi}
\end{aligned}
\end{equation}
Intuitively, Equation~\ref{eq:loss_cmi} seems to require two forward passes during training. However, our method avoids this entirely: since the model parameters $\theta$ are frozen and the term $\log p_{\theta}\big(y_t \mid \mathbf{X}, y_{<t}\big)$ is independent of the optimized video representation, it acts as a constant during training and can thus be safely removed. This elegantly demonstrates how our method further reduces training overhead. Finally, semantic consistency loss is defined as:
\begin{equation}
    \mathcal{L}_{\mathrm{S}}
    = -\sum_{t=1}^{L}
    \log p_{\theta}\big(y_t \mid \mathbf{Z}^*_v, \mathbf{X}, y_{<t}\big).
    \label{eq:final_loss}
\end{equation}

Interestingly, the final objective naturally reduces to a negative log-likelihood form. This does not imply that our method merely adopts the conventional language modeling loss. Rather, it reveals that the proposed conditional mutual information maximization objective can be transformed into a simple and tractable likelihood maximization problem. Therefore, the final loss can be viewed as returning to the fundamental principle of autoregressive generation: the optimized video representation should make the target response more probable under the frozen VideoLMM. Such a reduction is desirable, as it preserves the information-theoretic motivation of semantic consistency while avoiding additional complex estimators, auxiliary networks, or extra forward passes.

\subsection{Training and Inference}

To effectively parameterize the spatiotemporal-semantic aligner $\mathcal{M}$, we implement it as a lightweight bottleneck MLP. Formally, the structure of $\mathcal{M}$ is defined as:
\begin{equation}
\begin{aligned}
    \mathbf{U}_0
    &= \mathrm{GELU}
    \left(
    \mathrm{LN}(\mathbf{H}_v)\mathbf{W}_0^\top + \mathbf{b}_0
    \right), \\
    \mathbf{U}_r
    &= \mathrm{GELU}
    \left(
    \mathbf{U}_{r-1}\mathbf{W}_r^\top + \mathbf{b}_r
    \right), \\
    \mathcal{M}(\mathbf{H}_v)
    &= \alpha
    \left(
    \mathbf{U}_R\mathbf{W}_{R+1}^\top + \mathbf{b}_{R+1}
    \right),
\end{aligned}
\label{eq:aligner_structure}
\end{equation}
where $\mathrm{LN}(\cdot)$ denotes layer normalization, the second equation is recursively applied for $r=1,\ldots,R$, $R$ is the number of hidden layers in the bottleneck space, and $\alpha$ is a learnable scaling factor. The projection matrices satisfy $\mathbf{W}_0 \in \mathbb{R}^{d_b \times d}$, $\mathbf{W}_r \in \mathbb{R}^{d_b \times d_b}$, and $\mathbf{W}_{R+1} \in \mathbb{R}^{d \times d_b}$, where $d_b$ denotes the bottleneck dimension.

During training, the overall loss is defined as:
\begin{equation}
\label{eq:loss}
\mathcal{L}
= \mathcal{L}_{\mathrm{T}}
+ \lambda \mathcal{L}_{\mathrm{S}},
\end{equation}
where $\lambda$ is a hyperparameter that controls the relative weighting between the two loss terms in overall optimization. By jointly optimizing these two losses, we can benefit from enhancing the spatiotemporal consistency of video representations while preventing excessive spatiotemporal alignment from causing semantic deviation. This enables a collaborative optimization of both spatiotemporal and semantic consistency.

During inference, as shown on the right side of Figure~\ref{fig:overview}, the trained aligner is inserted after video representation extraction, allowing ViSSRes to mitigate hallucinations in VideoLMMs while introducing only negligible inference overhead.

\section{Experiments}

\begin{table*}
\centering
\small
\begin{tabular}{lccc|c|cccccc}
\toprule
\multirow{2}{*}{\textbf{Method}} & \multicolumn{3}{c|}{\textbf{VideoHallucer}} & \textbf{EventHallusion} & \multicolumn{5}{|c}{\textbf{VideoHallu}}\\
\cmidrule(lr){2-4} \cmidrule(lr){5-5} \cmidrule(lr){6-10}
 & Overall & Pct. Diff & FP Ratio & Overall & Align & S-T & Comm & Phys & Overall \\
\midrule
LLaVA-NeXT & 38.01 & 0.15 & 0.69 & 48.66 & 43.43 & 32.13 & 27.50 & 28.52 & 37.77 \\
+ TCD & 38.01 & 0.14 & 0.64 & 48.04 & 41.75 & 31.33 & 28.75 & 29.51 & 36.92 \\
+ MotionCD & 32.58 & 0.15 & 0.62 & 44.99 & 39.62 & 29.32 & 23.75 & 26.56 & 34.49 \\
+ DINO-HEAL & 14.65 & -0.18 & 0.26 & 48.90 & 42.54 & 31.73 & 25.00 & 27.54 & 36.85 \\
+ SmartSight & 19.10 & -0.26 & 0.36 & 62.10 & 39.26 & 35.60 & 31.02 & 31.57 & 36.29 \\
+ TA-AE & 32.64 & 0.14 & 0.70 & 67.00 & \textbf{53.91} & 36.12 & 40.74 & 38.22 & 45.66 \\
+ Ours & \textbf{38.32} & \textbf{0.05} & \textbf{0.54} & \textbf{68.46} & 52.08 & \textbf{39.36} & \textbf{50.00} & \textbf{39.34} & \textbf{47.34} \\
\midrule
Video-LLaVA & 16.80 & 0.36 & 0.91 & 46.70 & 45.90 & 37.75 & 18.75 & 31.80 & 40.33 \\
+ TCD & 17.21 & 0.37 & 0.92 & 45.23 & 45.01 & 37.35 & 23.75 & 33.11 & 40.26 \\
+ MotionCD & 24.18 & 0.22 & 0.71 & 40.34 & 38.50 & 30.92 & 20.00 & 29.84 & 34.56 \\
+ DINO-HEAL & 17.62 & 0.36 & 0.18 & 47.43 & 38.05 & 28.11 & 18.75 & 25.90 & 32.98 \\
+ SmartSight & 15.00 & 0.35 & 0.97 & 48.87 & 26.67 & 34.32 & 20.16 & 36.77 & 30.30 \\
+ TA-AE & 14.45 & 0.33 & 0.92 & 64.23 & \textbf{55.61} & 41.79 & 36.92 & 41.69 & 48.68 \\
+ Ours & \textbf{27.46} & \textbf{0.05} & \textbf{0.51} & \textbf{67.97} & 53.87 & \textbf{44.58} & \textbf{46.25} & \textbf{42.62} & \textbf{49.70} \\
\bottomrule
\end{tabular}
\caption{Comparison results on VideoHallucer, EventHallusion, and VideoHallu dataset. For the Pct.~Diff and FP~Ratio metrics, values closer to 0 and 0.5 respectively indicate better performance, while for the other metrics, larger values are better. Boldface indicates the best value.}
\label{tab:main_results}
\end{table*}

In this section, we evaluate ViSSRes's hallucination mitigation performance and general capability preservation. We further conduct ablation studies and qualitative analyses.
\subsection{Experimental Setup}
\paragraph{Models and Baselines} We adopt Video-LLaVA (8 frames)~\cite{lin2024video} and LLaVA-NeXT-Video (16 frames)~\cite{zhang2024llavanext-video} as backbones. In addition to the vanilla VideoLMMs, we compare our ViSSRes with SOTA inference-time intervention methods, including TCD~\cite{zhang2024eventhallusion}, MotionCD~\cite{kong2025mhbench}, DINO-HEAL~\cite{li2025vidhalluc}, SmartSight~\cite{sun2026smartsight}, and TA-AE~\cite{cai2026mitigating}. 

\paragraph{Evaluations}
We adopt VideoHallucer~\cite{wang2024videohallucer}, EventHallusion~\cite{zhang2024eventhallusion}, and VideoHallu~\cite{li2026videohallu} to evaluate the performance of hallucination mitigation. In addition, ActivityNet-QA~\cite{yu2019activitynet} and MMVU~\cite{Zhao_2025_CVPR} are used to assess general ability. Details of the evaluation procedure and benchmarks are provided in Appendix~\ref{sec:evaluation}.

\paragraph{Implementation Details} We randomly sample 3,000 instances from ShareGPT4Video~\cite{chen2024sharegpt4video} to train the aligner using the AdamW optimizer~\citep{loshchilov2017decoupled}. Following~\citet{jabri2020space}, we set the hyperparameter $\tau$ in Equation~\ref{eq:transition_matrix} to 0.07. The coefficient $\lambda$ in Equation~\ref{eq:loss} is set to 5. Additional details are provided in Appendix~\ref{sec:implementation}.

\subsection{Main Results}
\paragraph{Performance of Hallucination Mitigation}
As shown in Table~\ref{tab:main_results}, our method consistently achieves the best overall performance across VideoHallucer, EventHallusion, and VideoHallu on both backbone models. On VideoHallucer, our method not only obtains the highest Overall scores but also brings the Pct.~Diff closer to 0 and the FP~Ratio closer to 0.5, indicating more balanced hallucination mitigation. On EventHallusion, our method outperforms all baselines, including the strong TA-AE baseline, demonstrating its effectiveness in mitigating event-level hallucinations. On VideoHallu, our method achieves the best Overall scores and performs particularly well on the S-T, Comm, and Phys categories, although TA-AE is slightly better on Align. Overall, these results show that ViSSRes provides robust hallucination mitigation.

\paragraph{Performance of General Video Understanding}
Since our method focuses on hallucination mitigation, we further evaluate whether it affects the general video understanding capability of the backbone models. As shown in Table~\ref{tab:eval_general}, our method does not degrade performance on ActivityNet-QA or MMVU. Instead, it brings consistent improvements across both LLaVA-NeXT and Video-LLaVA. For our method slightly improves ActivityNet-QA Accuracy and Score, and notably increases the MMVU CoT score from 30.50 to 36.10. For Video-LLaVA, all metrics are also improved. These results demonstrate that our method can mitigate video hallucinations while preserving, and even slightly enhancing, the general video understanding capability of VideoLMMs.

\begin{table}
\small
\centering
\label{tab:three_line_short}
\begin{tabular}{lcc|cc}
\toprule
\multirow{2}{*}{Method}
& \multicolumn{2}{c|}{ActivityNet-QA}
& \multicolumn{2}{c}{MMVU} \\
\cmidrule(lr){2-3}\cmidrule(lr){4-5}
& Accuracy & Score & Direct & CoT\\
\midrule
LLaVA-NeXT &53.28 &3.43 &31.30 &30.50\\
+ Ours &53.86 &3.47 &32.00 &36.10\\
\midrule
Video-LLaVA &40.89 &2.95 &29.00 &28.20 \\
+ Ours &42.75 &3.01 &31.20 &29.60 \\
\bottomrule
\end{tabular}
\caption{Comparison results on ActivityNet-QA and MMVU dataset. Higher values indicate better performance across all metrics.}
\label{tab:eval_general}
\end{table}

\subsection{Ablation Study}

\paragraph{Role of Spatiotemporal and Semantic Consistency Alignment}
As shown in Figure~\ref{fig:motivation}, using only the spatiotemporal consistency alignment loss may bias video representations toward spatiotemporal consistency while deviating from the semantic space of VideoLMMs, resulting in performance degradation. This demonstrates the effectiveness of the proposed semantic consistency alignment. Moreover, Table~\ref{tab:ablation} shows that the overall performance first increases and then gradually decreases as $\lambda$ grows. Removing the spatiotemporal consistency loss leads to a clear performance drop, suggesting that semantic alignment alone is insufficient to preserve the temporal dynamics and spatial relationships of video features. These results indicate that neither semantic alignment nor spatiotemporal consistency alone can achieve optimal performance; instead, their synergy is essential for improving the model.

\begin{table}
\small
\centering
\begin{tabular}{lccc}
\toprule
Settings & Overall & Pct. Diff & FP Ratio \\
\midrule
w $\mathcal{L}_{\mathrm{T}}$, w $\mathcal{L}_{\mathrm{S}}$, $\lambda=0.1$ & 13.11 & 0.35 & 0.86 \\
w $\mathcal{L}_{\mathrm{T}}$, w $\mathcal{L}_{\mathrm{S}}$, $\lambda=1$ & 18.34 & 0.31 & 0.84 \\
w $\mathcal{L}_{\mathrm{T}}$, w $\mathcal{L}_{\mathrm{S}}$, $\lambda=5$ & 28.28 & 0.05 & 0.51 \\
w $\mathcal{L}_{\mathrm{T}}$, w $\mathcal{L}_{\mathrm{S}}$, $\lambda=10$ & 28.48 & 0.07 & 0.54 \\
w $\mathcal{L}_{\mathrm{T}}$, w $\mathcal{L}_{\mathrm{S}}$, $\lambda=50$ & 26.43 & 0.22 & 0.75 \\
w $\mathcal{L}_{\mathrm{T}}$, w $\mathcal{L}_{\mathrm{S}}$, $\lambda=100$ & 24.90 & 0.28 & 0.83 \\
w/o $\mathcal{L}_{\mathrm{T}}$, w $\mathcal{L}_{\mathrm{S}}$, $\lambda=5$ & 24.39 & 0.28 & 0.83\\
\bottomrule
\end{tabular}
\caption{Ablation results under different settings on the VideoHallucer using Video-LLaVA as the backbone.}
\label{tab:ablation}
\end{table}

\begin{figure}
  \raggedright
  \includegraphics[width=0.91\linewidth]{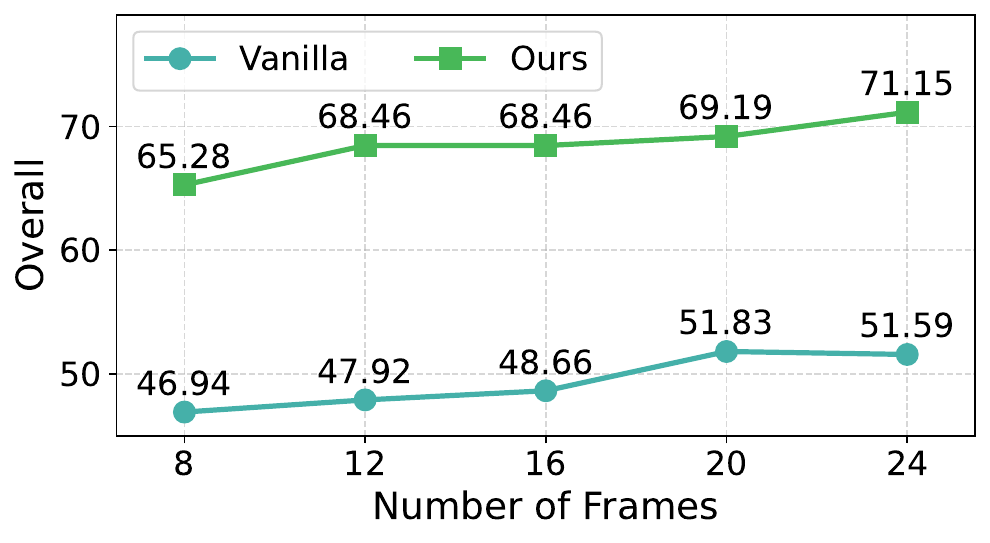}
  \caption{Performance comparison of Vanilla and Our method (LLaVA-NeXT-Video as backbone) across different numbers of frames on EventHallusion dataset.}
  \label{fig:frame_res}
\end{figure}

\paragraph{Impact of Number of Frames}
We adopt LLaVA-NeXT-Video as the backbone to investigate the robustness of ViSSRes to different numbers of frames on EventHallusion. Experiments are conducted on 8–24 frames with an interval of 4, and the results are shown in Figure~\ref{fig:frame_res}. While the vanilla model generally exhibits an upward trend as the number of frames increases, our method further improves performance across all frame settings. This demonstrates that ViSSRes remains effective under different input frame numbers and provides complementary gains beyond simply using more frames.

\paragraph{Impact of Training Data}
To verify whether ViSSRes is overly influenced by the training data, we train the aligner on the TA-AE~\cite{cai2026mitigating} and the LLaVA-Video-178K datasets~\cite{zhang2024llava}. More information is provided in Appendix~\ref{sec:implementation}. As shown in Table~\ref{tab:dataset_comparison}, training the aligner on different datasets consistently outperforms the vanilla model across all three benchmarks. ShareGPT4Video achieves the best V.Hallu score, while LLaVA-Video-178K obtains the highest V.Hallucer score and TA-AE Dataset performs best on MMVU. These results suggest that ViSSRes is robust to training data variations and benefits from diverse video-text supervision, rather than relying on dataset-specific overfitting.

\begin{table}
\centering
\small
\begin{tabular}{lccc}
\toprule
Training Dataset & V.Hallucer & V.Hallu & MMVU \\
\midrule
Vanilla & 16.80 & 40.33 & 29.00 \\
ShareGPT4Video & 28.28 & 49.70 & 31.20 \\
TA-AE Dataset & 27.77 & 47.41 & 40.80 \\
LLaVA-Video-178k & 28.59 & 47.93 & 38.08 \\
\bottomrule
\end{tabular}
\caption{Performance comparison of different training datasets. V.Hallucer and V.Hallu report the overall scores of VideoHallucer and VideoHallu, while MMVU uses the direct multiple-choice subset.}
\label{tab:dataset_comparison}
\end{table}

\subsection{Case Study}
\begin{figure}
  \centering
  \includegraphics[width=\linewidth]{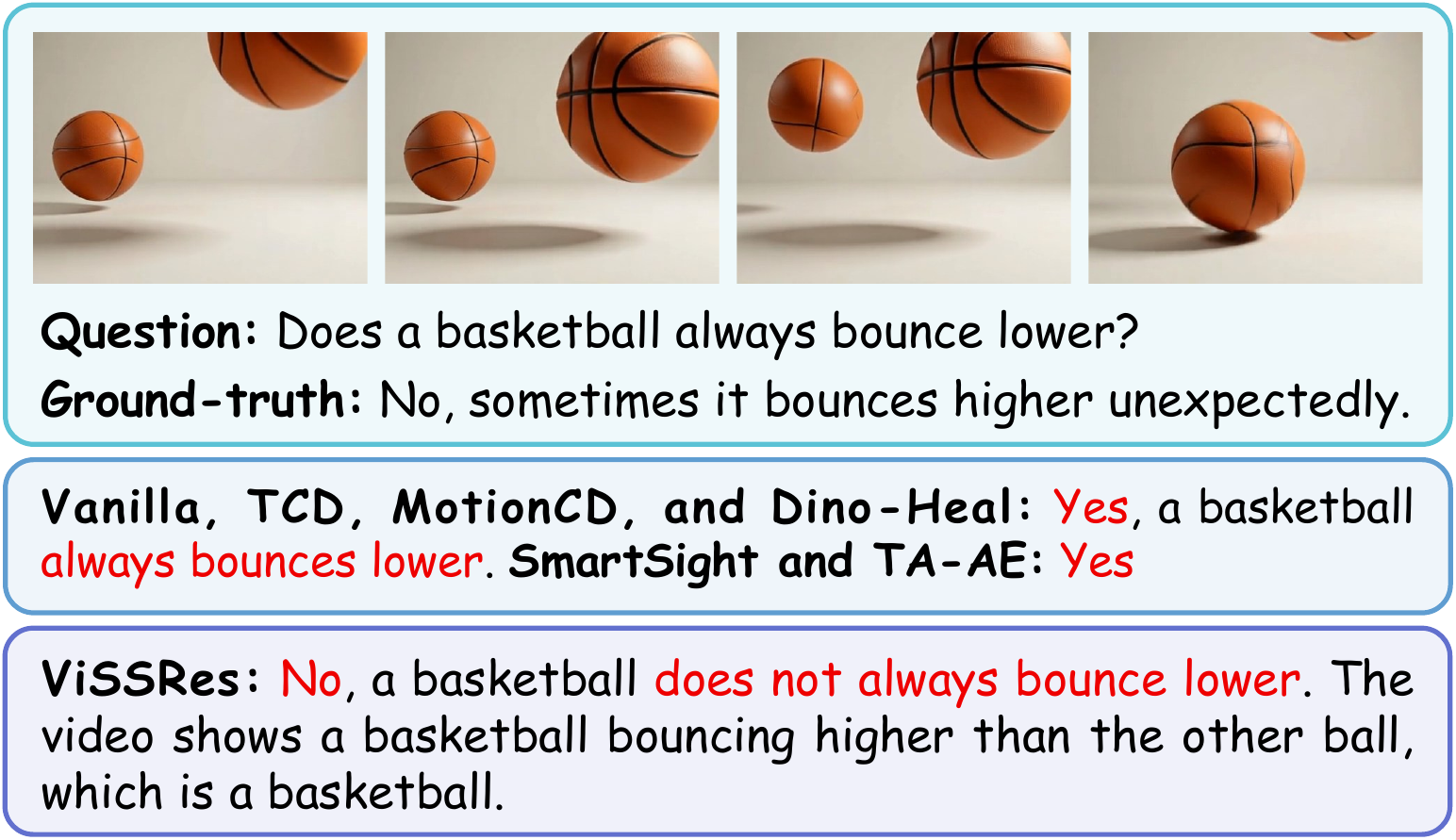}
  \caption{A representative case from the VideoHallu dataset using LLaVA-NeXT-Video as the backbone.}
  \label{fig:main_case}
\end{figure}

To further provide an intuitive analysis of the effectiveness of ViSSRes, we present a representative case in Figure~\ref{fig:main_case}. This case requires the model to determine whether a basketball always bounces lower, where the correct answer is "No". However, all compared methods answer "Yes", indicating that they tend to rely on commonsense priors or static object-level semantics rather than the actual motion evidence. In contrast, ViSSRes correctly answers "No" and identifies that one basketball bounces higher than the other. Since both objects are basketballs, this result cannot be obtained by object recognition alone; instead, it requires comparing their temporal trajectories and relative bouncing heights. This demonstrates that ViSSRes better captures fine-grained spatiotemporal dynamics and reduces errors caused by prior-biased reasoning. More cases are presented in Appendix~\ref{sec:more_cases}.

\section{Conclusion}
In this paper, we propose ViSSRes, an inference-time intervention method for mitigating hallucinations in VideoLMMs by optimizing video representations. We first show that directly enhancing spatiotemporal consistency via contrastive random walk is insufficient; excessive spatiotemporal alignment can even push video representations away from the model's language-semantic space, thereby impairing visual-to-language inference. Motivated by this observation, we introduce semantic consistency alignment and formulate a tractable training objective by maximizing the conditional mutual information between video representations and model responses given the question. 

Extensive experiments demonstrate that ViSSRes outperforms existing SOTA inference-time intervention methods, reducing hallucinations while preserving general video understanding ability. Ablation studies confirm the robustness of ViSSRes to different frame-number settings and training data variations. They also show that spatiotemporal and semantic consistency are complementary rather than interchangeable: only their joint optimization can reliably preserve both the spatiotemporal reliability and semantic decodability of video representations. Overall, ViSSRes highlights that hallucination mitigation requires not only more consistent visual representations, but also representations that remain semantically aligned with the response in a form interpretable to the language model.

\section*{Limitations}

Although ViSSRes mitigates hallucinations while preserving general video understanding ability, the current study is still limited in the exploration of the aligner architecture. ViSSRes adopts a lightweight MLP-style network as the spatiotemporal-semantic aligner, which keeps the method simple and efficient and introduces only negligible inference overhead. However, we do not systematically investigate other possible aligner designs. Therefore, the current results mainly demonstrate the effectiveness of a simple representation-level optimization, while whether more expressive or adaptive aligner architectures can further improve performance remains underexplored.


\bibliography{custom}

\appendix
\section{Evaluation and Benchmarks}
\label{sec:evaluation}

\paragraph{VideoHallucer Evaluation} VideoHallucer comprises both basic questions and hallucination-oriented questions, enabling not only the assessment of hallucination severity but also the analysis of model biases. Specifically, VideoHallucer evaluates multiple types of hallucinations, including Object-Relation Hallucination (ORH), Temporal Hallucination (TH), Semantic Detail Hallucination (SDH), Extrinsic Factual Hallucination (EFH), and Extrinsic Non-factual Hallucination (ENFH). For each video, a prediction is considered correct only if the model answers both the corresponding basic question and the hallucination question correctly, based on which the overall accuracy is computed. In addition, VideoHallucer reveals model biases by measuring the Yes Percentage Difference (Pct. Diff) and the False Positive Ratio (FP Ratio). Notably, all evaluation metrics are computed without adopting an LLM-as-Judge paradigm.

\paragraph{EventHallusion Evaluation} EventHallusion is a benchmark for diagnosing event-level hallucinations in VideoLMMs, focusing on failures in understanding temporally unfolding events and action sequences. It comprises 400 videos and 711 human-annotated questions across diverse domains, including daily life, sports, transportation, food, animals, and natural scenes. During evaluation, the selected deterministic subtasks compute accuracy solely based on binary labels, without relying on an LLM-as-Judge, thereby ensuring objective and consistent evaluation.

\paragraph{VideoHallu Evaluation} VideoHallu is a benchmark for evaluating and mitigating multi-modal hallucinations in synthetic video understanding. It organizes questions into four categories: Alignment (Align), Spatial-Temporal Consistency (S-T), Common Sense Reasoning (Comm), and Physics (Phys), covering hallucination patterns driven by language priors rather than visual evidence. Evaluation is conducted using GPT-4o-mini as an LLM-as-a-Judge to assess the consistency between model outputs and human-annotated ground truth.

\paragraph{ActivityNet-QA Evaluation} ActivityNet-QA is a classic benchmark dataset for video question answering research, focusing on evaluating models' abilities in temporal understanding and semantic reasoning over videos. Following~\citet{maaz2024video}, we conduct zero-shot question answering evaluation on this benchmark using GPT-3.5-Turbo with the same prompt template, and report Accuracy and Score.

\paragraph{MMVU Evaluation} MMVU is a multi-disciplinary benchmark for expert-level video understanding, evaluating multimodal large models on knowledge-intensive videos from specialized domains. It contains both multiple-choice and open-ended questions, and supports two evaluation settings: Direct answer generation and Chain-of-Thought (CoT) reasoning. For multiple-choice questions under the Direct setting, evaluation does not rely on an LLM-as-Judge, while all other settings are evaluated using GPT-4o.

\section{Implementation Details}
\label{sec:implementation}
Experiments are conducted on four NVIDIA A100 GPUs with 40GB memory. We train the aligner for one epoch with a batch size of 1, using a constant learning rate of $5 \times 10^{-6}$. During inference, the maximum generation length is set to 512 tokens, and greedy decoding is adopted as the default decoding strategy to ensure stability and reproducibility. For all baselines, we follow the parameter settings reported in their original papers to ensure a fair comparison.

For the experiments evaluating the impact of training data on ViSSRes, we randomly sample 3,000 instances from ShareGPT4Video and LLaVA-Video-178K, respectively, and train the aligner for one epoch. Since the TA-AE dataset contains only 996 instances, we train on TA-AE for three epochs to keep the effective training scale as consistent as possible. All other training settings are kept unchanged.

\begin{figure}[!t]
  \centering
  \includegraphics[width=0.9\linewidth]{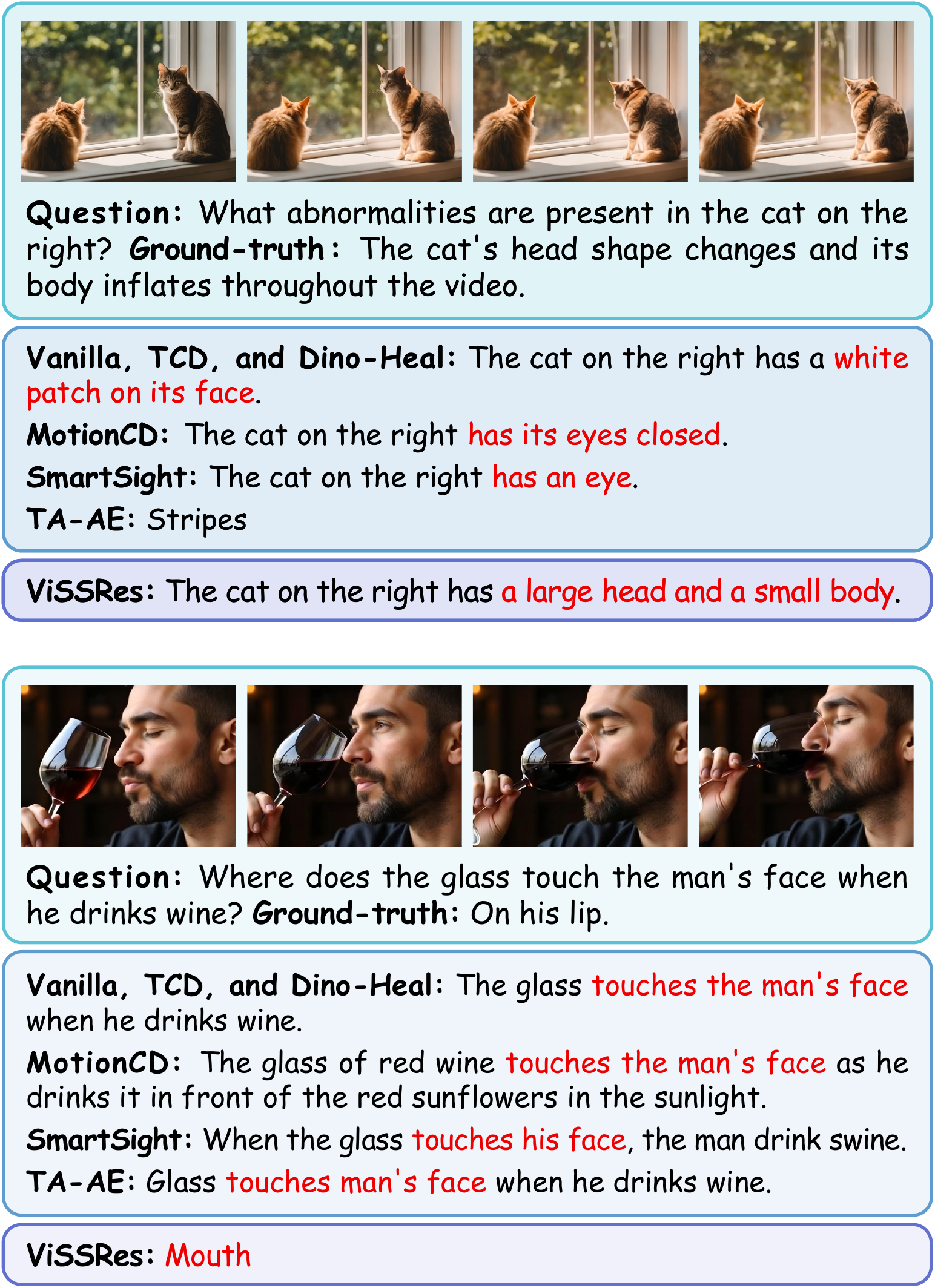}
  \caption{Representative cases from the VideoHallu dataset using Video-LLaVA as the backbone.}
  \label{fig:case1_2}
\end{figure}

\begin{figure}[!t]
  \centering
  \includegraphics[width=0.9\linewidth]{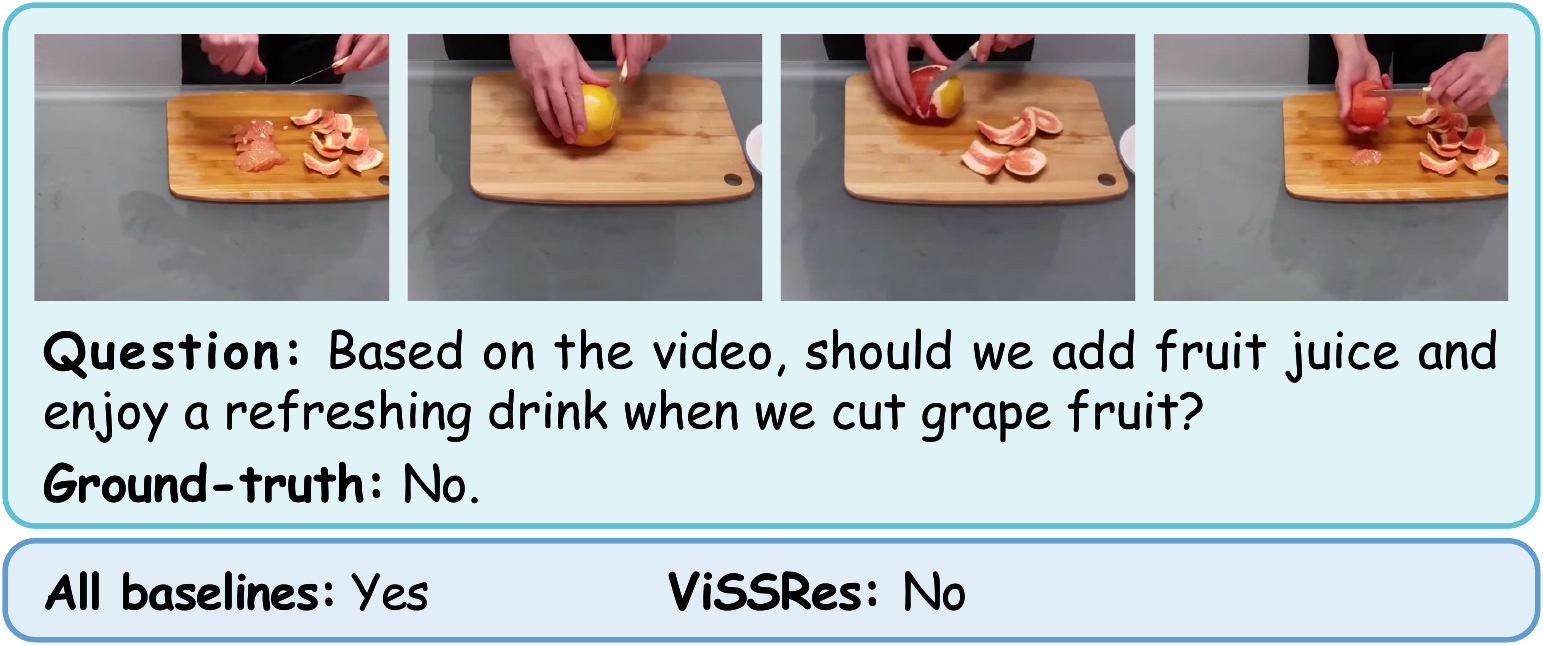}
  \caption{A representative case from the VideoHallucer dataset using LLaVA-NeXT-Video as the backbone.}
  \label{fig:case3}
\end{figure}

\begin{figure}[!t]
  \centering
  \includegraphics[width=0.9\linewidth]{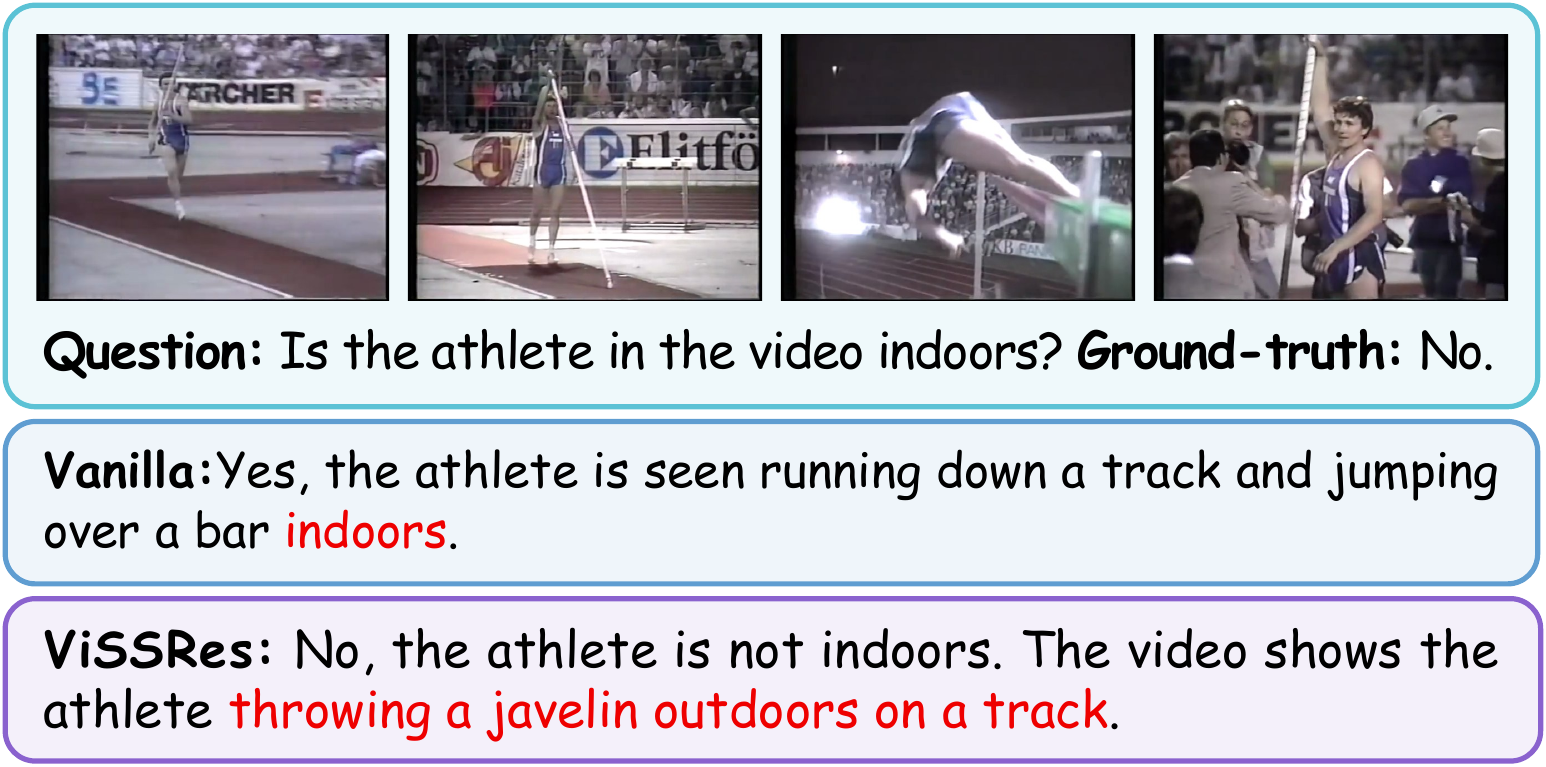}
  \caption{A representative case from the ActivityNet-QA dataset using Video-LLaVA as the backbone.}
  \label{fig:case4}
\end{figure}

\section{Representative Examples}
\label{sec:more_cases}
\begin{figure*}
  \centering
  \includegraphics[width=\linewidth]{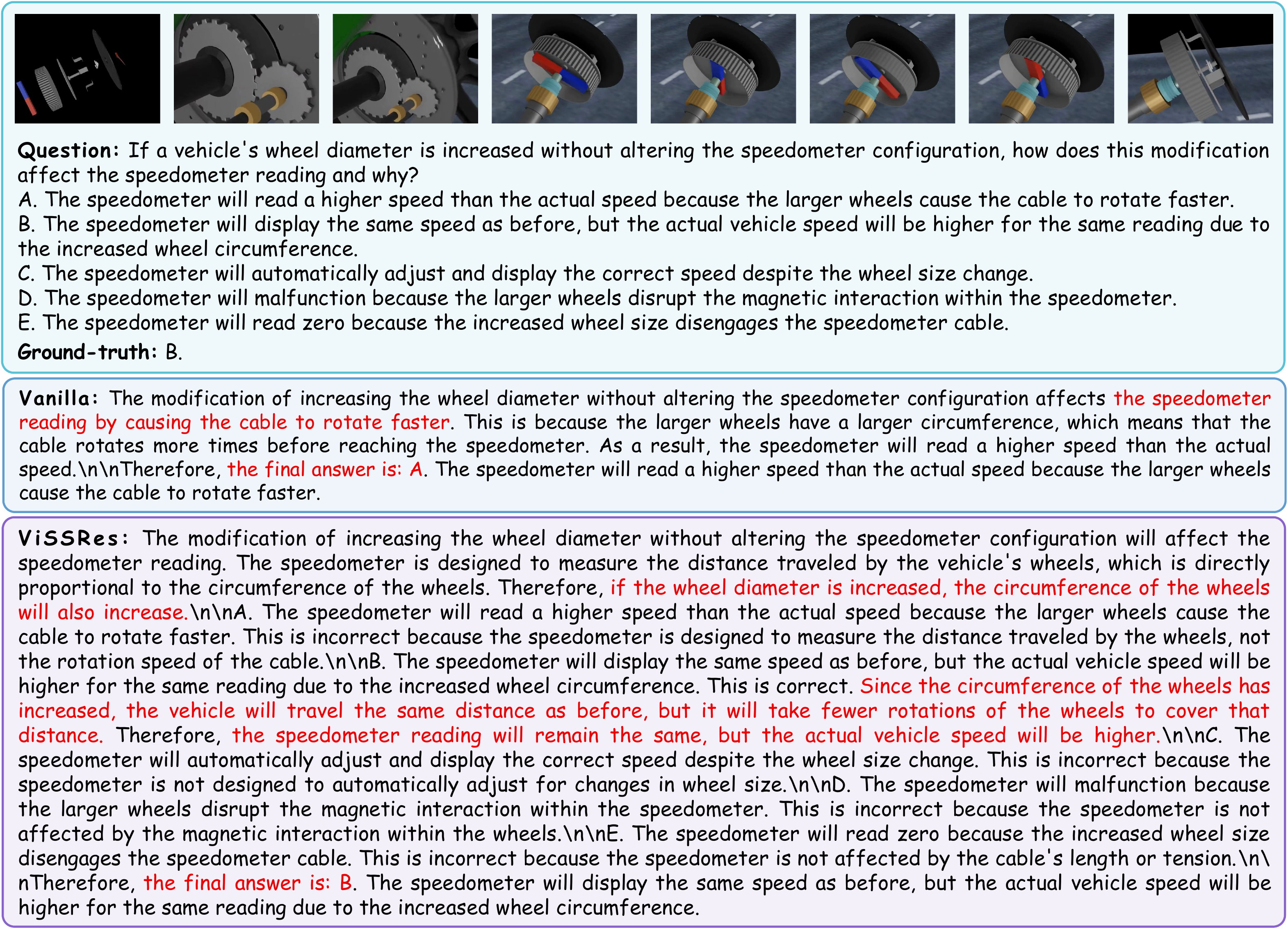}
  \caption{A representative case from the MMVU dataset using Video-LLaVA as the backbone.}
  \label{fig:case5}
\end{figure*}

To provide an intuitive illustration of the proposed method, we select several representative examples from different datasets and present them in Figures~\ref{fig:case1_2},~\ref{fig:case3},~\ref{fig:case4}, and~\ref{fig:case5}. These qualitative results further demonstrate the effectiveness of our method.

\end{document}